\documentclass[11pt,a4paper]{article}
\usepackage[nohyperref]{naaclhlt2018}
\usepackage{times}
\usepackage{latexsym}
\usepackage{url}
\usepackage{todonotes}
\usepackage{multirow}
\usepackage{xcolor}
\usepackage{times}
\usepackage{latexsym}
\usepackage{bbm}
\usepackage{amssymb}
\usepackage{amsmath}
\usepackage{amsfonts}
\usepackage{url}
\usepackage{multirow}
\usepackage{graphicx}
\aclfinalcopy 

\usepackage{amsmath,amssymb,amsthm}

\newcommand{\R}{\ensuremath{\mathbb{R}}}

\newcommand{\ra}{\ensuremath{\rightarrow}}

\newcommand{\paren}[1]{\left(#1\right)}

\newcommand{\abs}[1]{\left|#1\right|}

\usepackage{comment}
\usepackage{graphics}

\title{A Scalable Neural Shortlisting-Reranking Approach for Large-Scale Domain Classification in Natural Language Understanding}

\author{
  {\bf Young-Bum Kim} \hspace{10mm}
  {\bf Dongchan Kim} \hspace{10mm}
  {\bf Joo-Kyung Kim} \hspace{10mm}
  {\bf Ruhi Sarikaya} \\ 
  Amazon Alexa \\
  {\tt \{youngbum, dongchan, jookyk, rsarikay\}@amazon.com} 
}

\date{}

\begin{document}
\maketitle

\begin{abstract}

Intelligent personal digital assistants (IPDAs), a popular real-life application with spoken language understanding capabilities, can cover potentially thousands of overlapping domains for natural language understanding, and the task of finding the best domain to handle an utterance becomes a challenging problem on a large scale. In this paper, we propose a set of efficient and scalable neural shortlisting-reranking models for large-scale domain classification in IPDAs. The shortlisting stage focuses on efficiently trimming all domains down to a list of $k$-best candidate domains, and the reranking stage performs a list-wise reranking of the initial $k$-best domains with additional contextual information. We show the effectiveness of our approach with extensive experiments on 1,500 IPDA domains.


\end{abstract}

\section{Introduction}

Natural language understanding (NLU) is a core component of intelligent personal digital assistants (IPDAs) such as Amazon Alexa, Google Assistant, Apple Siri, and Microsoft Cortana~\cite{sarikaya-IEEE-SPM-paper}. A well-established approach in current real-time systems is to classify an utterance into a domain, followed by domain-specific intent classification and slot sequence tagging~\cite{Tur2011}. A domain is typically defined in terms of a specific application or a functionality such as weather, calendar and music, which narrows down the scope of NLU for a given utterance. A domain can also be defined as a collection of relevant intents; assuming an utterance belongs to the calendar domain, possible intents could be to create a meeting or cancel one, and possible extracted slots could be people names, meeting title and date from the utterance. Traditional IPDAs cover only tens of domains that share a common schema. The schema is designed to separate out the domains in an effort to minimize language ambiguity. A shared schema, while addressing domain ambiguity, becomes a bottleneck as new domains and intents are added to cover new scenarios. Redefining the domain, intent and slot boundaries requires relabeling of the underlying data, which is very costly and time-consuming. On the other hand, when thousands of domains evolve independently without a shared schema, finding the most relevant domain to handle an utterance among thousands of overlapping domains emerges as a key challenge.

The difficulty of solving this problem at scale has led to stopgap solutions, such as requiring an utterance to explicitly mention a domain name and restricting the expression to be in a predefined form as in ``Ask \texttt{ALLRECIPES}, how can I bake an apple pie?'' However, such solutions lead to an unintuitive and unnatural way of conversing and create interaction friction for the end users. For the example utterance, a more natural way of saying it is simply, ``How can I bake an apple pie?'' but the most relevant domain to handle it now becomes ambiguous. There could be a number of candidate domains and even multiple overlapping recipe-related domains that could handle it.

In this paper, we propose efficient and scalable shortlisting-reranking neural models in two steps for effective large-scale domain classification in IPDAs. The first step uses light-weight BiLSTM models that leverage only the character and word-level information to efficiently find the $k$-best list of most likely domains. The second step uses rich contextual information later in the pipeline and applies another BiLSTM model to a list-wise ranking task to further rerank the $k$-best domains to find the most relevant one. We show the effectiveness of our approach for large-scale domain classification with an extensive set of experiments on 1,500 IPDA domains. 

\section{Related Work}

Reranking approaches attempt to improve upon an initial ranking by considering additional contextual information. Initial model outputs are trimmed down to a subset of most likely candidates, and each candidate is combined with additional features to form a hypothesis to be re-scored. Reranking has been applied to various natural language processing tasks, including machine translation~\cite{shen2004discriminative}, parsing~\cite{collins2005discriminative}, sentence boundary detection~\cite{roark2006reranking}, named entity recognition~\cite{nguyen2010kernel}, and supertagging~\cite{chen2002reranking}.

In the context of NLU or SLU systems, \newcite{Morbini2012} showed a reranking approach using $k$-best lists from multiple automatic speech recognition (ASR) engines to improve response category classification for virtual museum guides. \newcite{Basili2013} showed that reranking multiple ASR candidates by analyzing their syntactic properties can improve spoken command understanding in human-robot interaction, but with more focus on ASR improvement. \newcite{xu2014contextual} showed that multi-turn contextual information and recurrent neural networks can improve domain classification in a multi-domain and multi-turn NLU system. There have been many other pieces of prior work on improving NLU systems with pre-training~\cite{kim2015pre,celikyilmaz2016new,kim2017pre}, multi-task learning~\cite{zhang2016joint,Liu+2016,kim2017onenet}, transfer learning~\cite{el2014extending,kim2015weakly,kim2015new,chen2016zero,yang2017transfer}, domain adaptation~\cite{kim2016frustratingly, jaech2016domain, liu2017multi, kim2017domain, kim2017advr} and contextual signals~\cite{Bhargava2013EasyCI,chen2016end, hori2016context, kim2017speaker}.

To our knowledge, the work by \newcite{Robichaud2014,Crook2015,Khan2015} is most closely related to this paper. Their approach is to first run a complete pass of all 3 NLU models of binary domain classification, multi-class intent classification, and sequence tagging of slots across all domains. Then, a hypothesis is formed per domain using the semantic information provided by the domain-intent-slot outputs as well as many other contextual and cross-hypothesis features such as the presence of a slot tagging type in any other hypotheses. Reranking the hypotheses with Gradient Boosted Decision Trees~\cite{Friedman2001,Burges2011} has been shown to improve domain classification performance compared to using only domain classifiers without reranking.


Their approach however suffers from the following two limitations. First, it requires running all domain-intent-slot models in parallel across all domains. Their work considers only 8 or 9 distinct domains, and the approach has serious practical scaling issues when the number of domains scales to thousands. Second, contextual information, especially cross-hypothesis features, that is crucial for reranking is manually designed at the feature level with a sparse representation.

Our work in this paper addresses both of these limitations with a scalable and efficient two-step shortlisting-reranking approach, which has a neural ranking model capturing cross-hypothesis features automatically. To our knowledge, this work is the first in the literature on large-scale domain classification for a real IPDA production system with a scale of thousands of domains. Our LSTM-based list-wise ranking approach also makes a novel contribution to the existing literature in the context of IPDA and NLU systems. In this work, we limit our scope to first-turn utterances and leave multi-turn conversations for future work.

\section{Shortlisting-Reranking Architecture}
\label{sec:architecture}
Our shortlisting-reranking models process an incoming utterance as follows. (1) \textit{Shortlister} performs a naive, fast ranking of all domains to find the $k$-best list using only the character and word-level information. The goal here is to achieve high domain recall with maximal efficiency and minimal information and latency.
(2) For each domain in the $k$-best list, we prepare a hypothesis per domain with additional contextual information, including domain-intent-slot semantic analysis, user preferences, and domain index of popularity and quality.
(3) A second ranker called \textit{Hypotheses Reranker (HypRank)} performs a list-wise ranking of the $k$ hypotheses to improve on the initial naive ranking and find the best hypothesis, thus domain, to handle the utterance.

\begin{figure*}[t!]
  \begin{center}
    \includegraphics[width=0.70\textwidth]{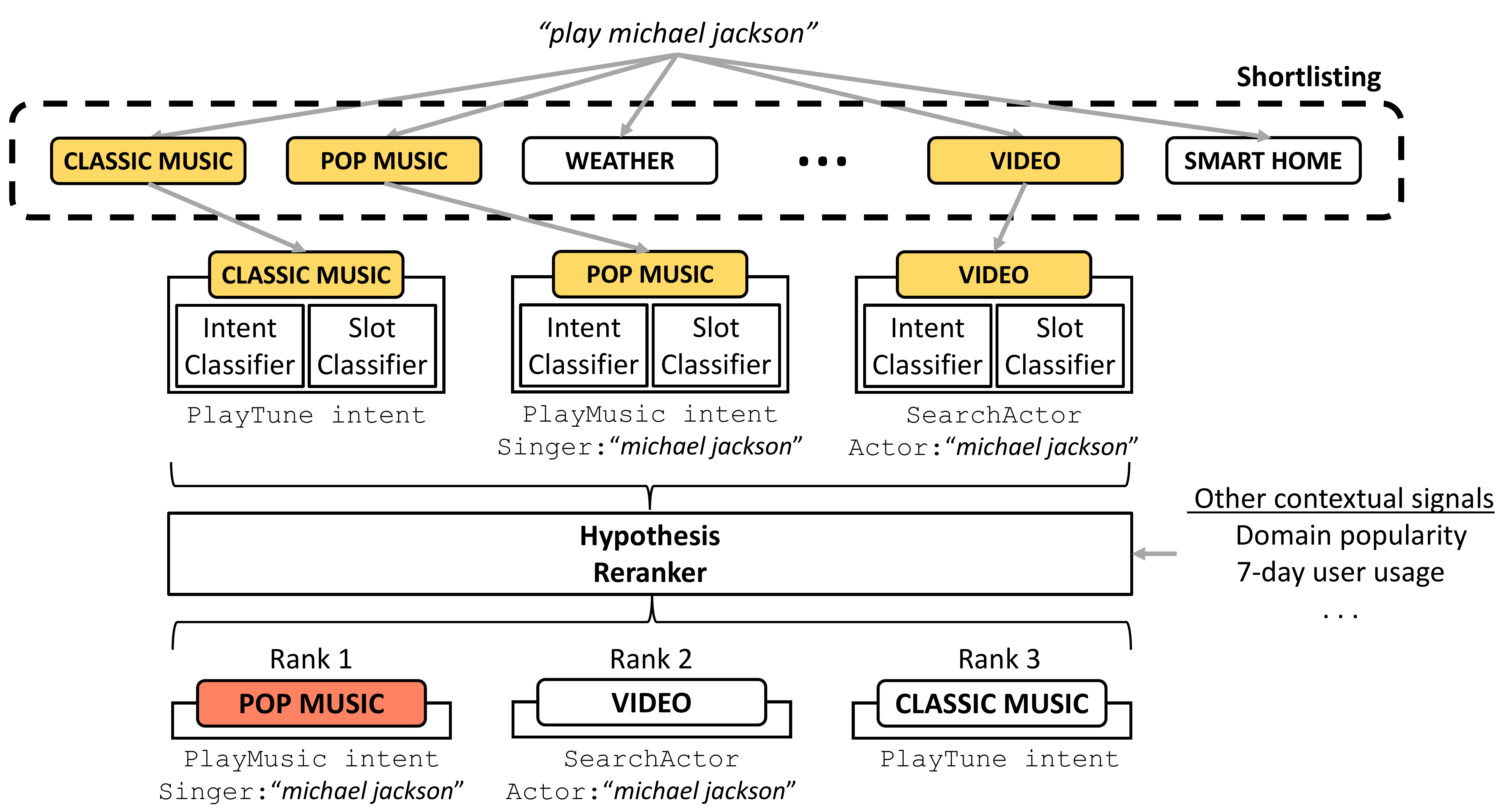}
  \end{center}
  \vspace{-0.6cm}
  \caption{\small A high-level flow of our two-step shortlisting-reranking approach given an utterance to an IPDA.}
  \label{fig:hyprank_overview}
\end{figure*}

Figure~\ref{fig:hyprank_overview} illustrates the steps with an example utterance, ``\textit{play michael jackson}.'' Based on character and word features, \textit{shortlister} returns the $k$-best list in the order of \texttt{CLASSIC MUSIC, POP MUSIC, and VIDEO} domains. \texttt{CLASSIC MUSIC} outputs \texttt{PlayTune} intent, but without any slots, low domain popularity, and no usage history for the user, its ranking is adjusted to be last. \texttt{POP MUSIC} outputs \texttt{PlayMusic} intent and \texttt{Singer} slot for ``\textit{michael jackson}'', and with frequent user usage history, it is determined to be the best domain to handle the utterance.

In our architecture, key focus is on efficiency and scalability. Running full domain-intent-slot semantic analysis for thousands of domains imposes a significant computational burden in terms of memory footprint, latency and number of machines, and it is impractical in real-time systems. For the same reason, this work only uses contextual information in the reranking stage, and the utility of including it in the shortlisting stage is left for future work.
\section{Shortlister}
\label{sec:shortlister}
\begin{figure}[t!]
  \begin{center}
    \includegraphics[width=0.73\columnwidth]{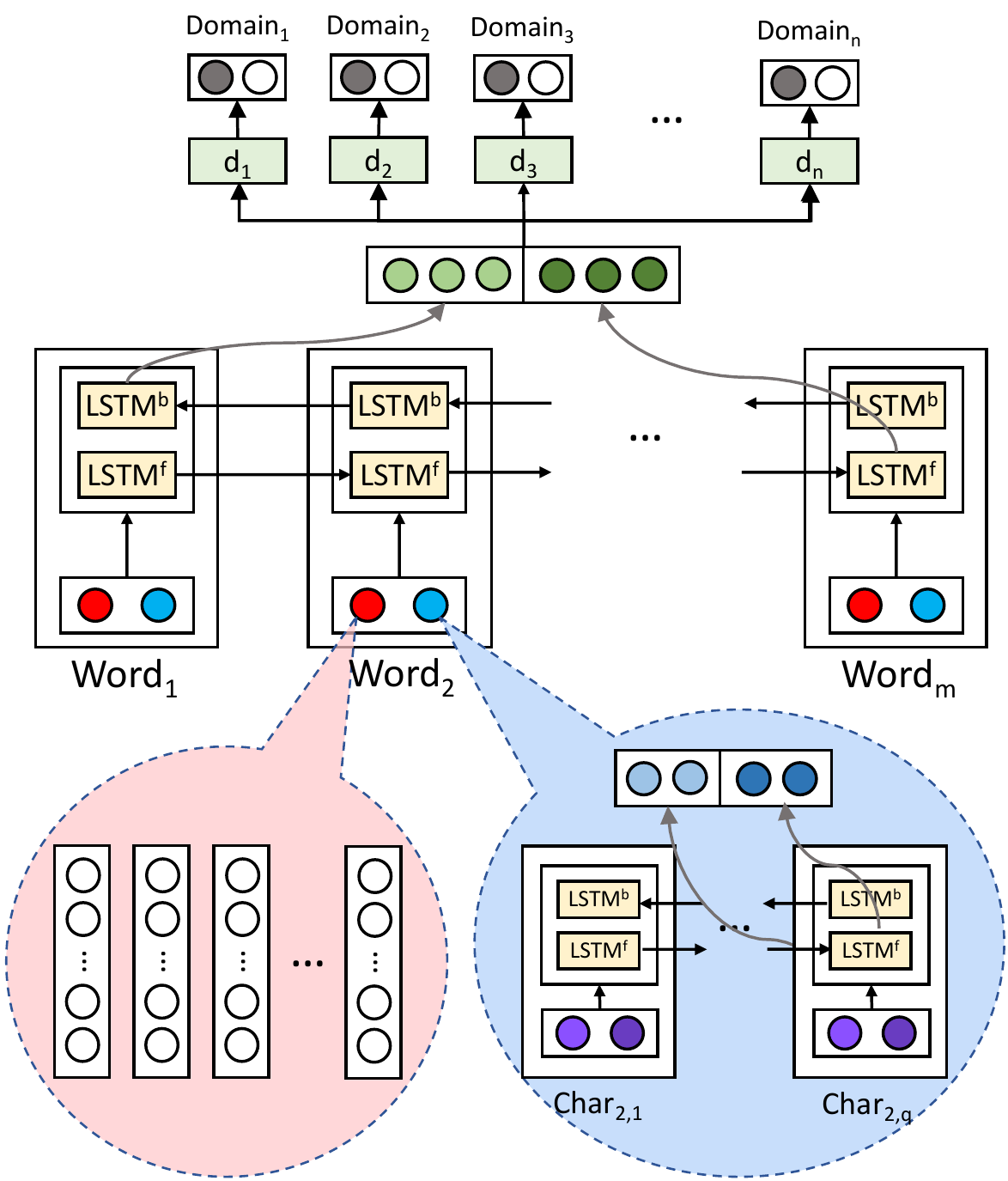}
  \end{center}
  \vspace{-0.55cm}
  \caption{\small The architecture of our neural shortlisting model that uses character and word-level information of a given utterance.}
  \label{fig:shortlisting_model}
\end{figure}

Shortlister consists of three layers: an orthography-sensitive character and word embedding layer, a BiLSTM layer that makes a vector representation for the words in a given utterance, and an output layer for domain classification. Figure~\ref{fig:shortlisting_model} shows the overall shortlister architecture.

\paragraph{Embedding layer}
In order to capture character-level patterns, we construct an orthography-sensitive word embedding layer \cite{Ling2015,Ballesteros2015}.
Let $\mathcal{C}$, $\mathcal{W}$, and $\oplus$ denote the set of characters, the set of words, and the vector concatenation operator, respectively.
We represent an LSTM as a mapping $\phi:\R^d \times \R^{d'} \ra \R^{d'}$
that takes an input vector $x$ and a state vector $h$ to output a new state vector $h' = \phi(x, h)$\footnote{We omit cell variable notations for simple LSTM formulations.}.
The model parameters associated with this layer are:
\begin{align*}
\textbf{Char embedding: } &e_c \in \R^{25} \text{ for each } c \in \mathcal{C} \\
\textbf{Char LSTMs: } &\phi^{\mathcal{C}}_f, \phi^{\mathcal{C}}_b: \R^{25} \times \R^{25} \ra \R^{25} \\
\textbf{Word embedding: } &e_w \in \R^{100} \text{ for each } w \in \mathcal{W}
\end{align*}

Let $\left(w_1, \dots, w_m\right)$ denote a word sequence where word $w_i\in\mathcal{W}$ has character $w_i(j) \in \mathcal{C}$ at position $j$.
This layer computes an orthography-sensitive word representation $v_i \in \R^{150}$ as:\footnote{We randomly initialize state vectors such as $f^{\mathcal{C}}_0$ and $b^{\mathcal{C}}_{\abs{w_i}+1}$.}
\begin{align*}
f^{\mathcal{C}}_j &= \phi^{\mathcal{C}}_f\paren{e_{w_i(j)}, f^{\mathcal{C}}_{j-1}} &&\forall j = 1 \ldots \abs{w_i} \\
b^{\mathcal{C}}_j &= \phi^{\mathcal{C}}_b\paren{e_{w_i(j)}, b^{\mathcal{C}}_{j+1}} &&\forall j = \abs{w_i} \ldots 1\\
v_i &= f^{\mathcal{C}}_{\abs{w_i}} \oplus b^{\mathcal{C}}_1 \oplus e_{w_i} &&
\end{align*}

\paragraph{BiLSTM layer}
We utilize a BiLSTM to encode the word vector sequence $(v_1, \dots, v_m)$. 
The BiLSTM outputs are generated as:
\begin{align*}
f^{\mathcal{W}}_i &= \phi^{\mathcal{W}}_f\paren{v_i, f^{\mathcal{W}}_{i-1}} &\forall i = 1 \ldots m& \\
b^{\mathcal{W}}_i &= \phi^{\mathcal{W}}_b\paren{v_i, b^{\mathcal{W}}_{i+1}} &\forall i = m \ldots 1&\quad
\end{align*}
where $\phi^{\mathcal{W}}_f, \phi^{\mathcal{W}}_f: \R^{150} \times \R^{100} \ra \R^{100}$ are the forward LSTM and the backward LSTM, respectively. An utterance representation $h \in \R^{200}$ is induced by concatenating the outputs of the both LSTMs as:
\begin{align*}
h &= f^{\mathcal{W}}_m \oplus b^{\mathcal{W}}_1
\end{align*}

\paragraph{Output layer}
\label{ssec:shortlister_output_layer}
We map the word LSTM output $h$ to a $n$-dimensional output vector with a linear transformation. Then, we take a softmax function either over the entire domains (\textit{$softmax_a$}) or over two classes (in-domain or out-of-domain) for each domain (\textit{$softmax_b$}).

\textit{$softmax_a$} is used to set the sum of the confidence scores over the entire domains to be 1. We can obtain the outputs as:
\begin{equation*}
o = softmax\left(W \cdot h + b\right)
\end{equation*}
where W and b are parameters for a linear transformation.

For training, we use cross-entropy loss, which is formulated as follows:
\begin{equation}
\mathcal{L}_{a} = -\sum_{i=1}^{n} l_i \log o_i
\end{equation}
where $l$ is a $n$-dimensional one-hot vector whose element corresponding to the position of the ground-truth hypothesis is set to 1.

\textit{$softmax_b$} is used to set the confidence score for each domain to be between 0 and 1.
While \textit{$softmax_a$} tends to highlight only the ground-truth domain while suppressing all the rest, \textit{$softmax_b$} is designed to produce a more balanced confidence score per domain independent of other domains. When using \textit{$softmax_b$}, we obtain a 2-dimensional output vector for each domain as follows:
\begin{equation*}
o^i = softmax\left(W^i \cdot h + b^i\right)
\end{equation*}
where $W^i$ is a 2 by 200 matrix and $b^i$ is a 2-dimensional vector; $o_1^i$ and $o_2^i$ denote the in-domain probability and the out-of-domain probability, respectively.
The loss function is formulated as follows:
\begin{equation}
\mathcal{L}_{b} = -\sum_{i=1}^{n} \left\{l_i \log o_1^i + \frac{1-l_i}{n-1} \log o_2^i\right\}
\end{equation}
where we divide the second term by $n-1$ so that $o_1^i$ and $o_2^i$ are balanced in terms of the ratio of the training examples for a domain to those for other domains.


\section{Hypotheses Reranker (HypRank)}
\label{sec:hyprank}

Hypotheses Reranker (HypRank) comprises of two components: hypothesis representation and a BiLSTM model for reranking a list of hypotheses. We use the term reranking since we improve upon the initial ranking from Shortlister's $k$-best list. In our problem context, a hypothesis is formed per domain with additional semantic and contextual information, and selecting the highest-scored hypothesis means selecting the domain represented in that hypothesis for final domain classification.

\begin{figure}[t]
  \begin{center}
    \includegraphics[width=1.0\columnwidth]{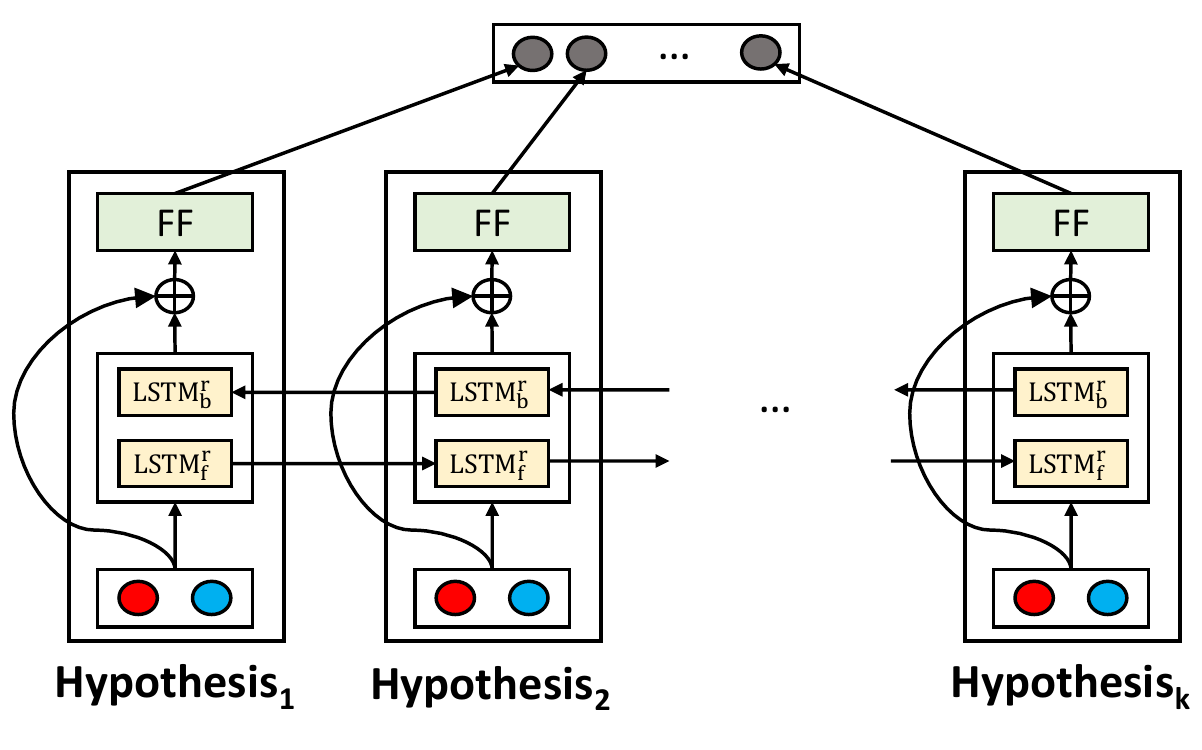}
  \end{center}
  \vspace{-0.6cm}
  \caption{\small The architecture of our neural Hypotheses Reranker model that takes in $k$ hypotheses with rich contextual information for more refined ranking.}
  \label{fig/expert}
\end{figure}

HypRank, illustrated in Figure \ref{fig/expert}, is a list-wise ranking approach in that it considers the entire list of hypotheses before giving a reranking score for each hypothesis. While previous work manually encoded cross-hypothesis information at the feature level~\cite{Robichaud2014,Crook2015,Khan2015}, our approach is to let a BiLSTM layer automatically capture that information and learn appropriate representations at the model level. In addition to giving detail of useful contextual signals for IPDAs, we also introduce the use of pre-trained domain, intent and slot embeddings in this section.

\subsection{Hypothesis Representation}
\label{ssec:input_feat}

A hypothesis is formed for each domain with the following three categories of contextual information: NLU interpretation, user preferences, and domain index.

\paragraph{NLU interpretation}
Each domain has three corresponding NLU models for binary domain classification, multi-class intent classification, and sequence tagging for slots. From the domain-intent-slot semantic analysis, 
we use the confidence score from the shortlister, the intent classification confidence, Viterbi path score of the slot sequence from a slot tagger, and the average confidence score of the tagged slots\footnote{We use off-the-shelf intent classifiers and slot taggers achieving 98\% and 96\% accuracies on average, respectively.}.



To pre-train domain embeddings, we use a word-level BiLSTM with each utterance as a sequence of word embedding vector $\in \R^{100}$ in the input layer. The BiLSTM outputs, each a vector $\in \R^{25}$, are concatenated and projected to an output vector for all domains in the output layer. The learned projection weight matrix is extracted as domain embeddings. The output vector dimension used was $\in \R^{1500}$ for the large-scale setting and $\in \R^{20}$ for the traditional small-scale setting in our experiments (Section~\ref{ssec:experimental_setup}). For intent and slot embeddings, we take the same process with the only difference in the output vector with the dimension $\in \R^{6991}$ for all unique intents across all domains and with the dimension $\in \R^{2845}$ for all unique slots.

Once pre-trained, the domain or intent embeddings are used simply as a lookup table per domain or per intent. For slot embeddings, there can be more than one slot per utterance, and in case of multiple tagged slots, we sum up each slot embedding vector to combine the information. In summary, these are the three domain-intent-slots embeddings we used: $e^{d} \in \R^{50}$ for a domain vector, $e^{i} \in \R^{50}$ for an intent vector, and $e^{s} \in \R^{50}$ for a vector of slots.



\paragraph{User Preferences} User-specific signals are designed to capture each user's behavioral history or preferences. In particular, we encode whether a user has specific domains enabled in his/her IPDA setting and whether he/she triggered certain domains within 7, 14 or 30 days in the past.

\paragraph{Domain Index} From this category, we encode domain popularity and quality as rated by the user population. For example, if the utterance ``\textit{I need a ride to work}'' can be equally handled by \texttt{TAXI\_A} domain or \texttt{TAXI\_B} domain but the user has never used any, the signals in this category could give a boost to \texttt{TAXI\_A} domain due to its higher popularity.

\subsection{HypRank Model}
\label{ssec:rerank_model}
The proposed model is trained to rerank the domain hypotheses formed from Shortlister results.
Let $\left(p_1, \dots, p_k\right)$ be the sequence of $k$ input hypothesis vectors that are sorted in decreasing order of Shortlister scores.

We utilize a BiLSTM layer for transforming the input sequence to the BiLSTM output sequence $\left(h_1, \dots, h_k\right)$ as follows:
\begin{align*}
f_i^r &= \phi_f^r\left(p_i, f_{i-1}^r\right)\\
b_i^r &= \phi_b^r\left(p_i, b_{i+1}^r\right)\\
h_i &= f_i^r \oplus b_i^r \qquad \forall i \in \left\{1, \dots, k\right\},
\end{align*}
where $\phi_f^r$ and $\phi_b^r$ are the forward LSTM and the backward LSTM, respectively.

Since the BiLSTM utilizes both the previous and the next sub-sequences as the context, each of the BiLSTM outputs is computed considering cross-hypothesis information.

For the $i$-th hypothesis, we either sum or concatenate the input vector and the BiLSTM output to utilize both of them as an intermediate representation as $g_i = d_i \oplus h_i$.
Then, we use a feed-forward neural network with a single hidden layer to transform $g$ to a $k$-dimensional vector $p$ as follows:
\begin{equation*}
p_i = W_2 \cdot \sigma\left(W_1 \cdot g_i + b_1\right) + b_2 \quad \forall i \in \left\{1, \dots, k\right\},
\end{equation*}

\begin{table}[t!]
\setlength\belowcaptionskip{-8pt}
\small
\centering
\begin{tabular}{rcl}
Category                 & $|\mathcal{D}|$  & Example                \\ \hline
\texttt{Device}          & 177              & smart home, smart car  \\
\texttt{Food}            & 99               & recipe, nutrition      \\
\texttt{Ent.}            & 465              & movie, music, game     \\
\texttt{Info.}           & 399              & travel, lifestyle      \\
\texttt{News}            & 159              & local, sports, finance \\
\texttt{Shopping}        & 39               & retail, food, media    \\
\texttt{Util.}           & 162              & productivity, weather  \\ \hline
Total                    & 1,500 \\\hline
\end{tabular}
\caption{\small The categories of the 1,500 domains for our large-scale IPDA. $|\mathcal{D}|$ denotes the number of domains.}
\label{tab:domain_stats}
\end{table}

where $\sigma$ indicates scaled exponential linear unit (SeLU) for normalized activation outputs \cite{Klambauer2017}; the outputs of all the hypotheses are generated by using the same parameter set $\{W_1, b_1, W_2, b_2\}$ for consistency regardless of the hypothesis order.

Finally, we obtain a $k$-dimensional output vector $o$ by taking a softmax function:
\begin{equation*}
o = softmax\left(p\right).
\end{equation*}

$argmax_i\{o_1, .., o_k\}$ is the index of the predicted hypothesis after the reranking.
Cross entropy is used for training as follows:
\begin{equation}
\mathcal{L}_r = -\sum_{i=1}^{k} l_i\log o_i,
\end{equation}
where $l$ is a $k$-dimensional ground-truth one-hot vector.

\section{Experiments}
This section gives detail of our experimental setup, followed by results and discussion.

\subsection{Experimental Setup}
\label{ssec:experimental_setup}

We evaluated our shortlisting-reranking approach in two different settings of traditional small-scale IPDA and large-scale IPDA for comparison:

\paragraph{Traditional IPDA} For this setting, we simulated the traditional small-scale IPDA with only 20 domains that are commonly present in any IPDAs. Since these domains are built-in, which are carefully designed to be non-overlapping and of high quality, the signals from user preferences and domain index become irrelevant compared to the large-scale setting.  The dataset comprises of more than 4M labeled utterances in text evenly distributed across 20+ domains.

\begin{table}[t!]
\setlength\belowcaptionskip{-10pt}
\small
\centering
\begin{tabular}{l|ccccc}
            & \begin{tabular}[c]{@{}l@{}}SL\\ train\end{tabular} & \begin{tabular}[c]{@{}l@{}}SL\\ dev\end{tabular} & \begin{tabular}[c]{@{}l@{}}HR\\ train\end{tabular} & \begin{tabular}[c]{@{}l@{}}HR\\ dev\end{tabular} & test \\ \hline
Traditional &  3M      & 415K   & 715K   & 20K   &  420K   \\
Large-Scale         &  5M      & 530K   & 830K   & 20K   &  530K   \\ \hline
\end{tabular}
\caption{\small The number of train, development and test utterances. SL denotes Shortlister and HR denotes HypRank.}
\label{tab:exam_nums}
\end{table}

\paragraph{Large-Scale IPDA} This setting is a large-scale IPDA with 1,500 domains as shown in Table~\ref{tab:domain_stats} that could be overlapping with a varying level of quality. For instance, there could be multiple domains to get a recipe, and a high quality domain could have more recipes with more capabilities such as making recommendations compared to a low quality one. The dataset comprises of more than 6M utterances having strict invocation patterns. For instance, we extract \textit{``get me a ride''} as a preprocessed sample belonging to \textit{TAXI} skill for the original utterance, \textit{``Ask \{TAXI\} to \{get me a ride\}.''}


\paragraph{Shortlister}
For Shortlister, we show the results of using 2 different softmax functions of \textit{$softmax_a$} (\textit{$smx_a$}) and \textit{$softmax_b$} (\textit{$smx_b$}) as described in Section~\ref{sec:shortlister}. The results are shown in $k$-best classification accuracies, where the 5-best accuracy means the percentage of test samples that have the ground-truth domain included in the top 5 domains returned by Shortlister.

\paragraph{Hypotheses Reranker}
We also evaluate different variations of the reranking model for comparison.

\begin{itemize}
\setlength{\itemsep}{2pt}
\setlength{\parskip}{1pt}
\item \textit{$SL$}: Shortlister $1$-best result, which is our baseline without using a reranking model.
\item \textit{$LR$}: LR point-wise: A binary logistic regression model with the hypothesis vector as features (see Section~\ref{ssec:input_feat}). We run it for each hypothesis made from Shortlister's $k$-best list and select the highest-scoring one, hence the domain in that hypothesis.
\item \textit{$N^{PO}$}: Neural point-wise: A feed-forward (FF) layer between the hypothesis vector and the nonlinear output layer. We run it for each hypothesis made from Shortlister's $k$-best list and select the highest-scoring hypothesis.
\item \textit{$N^{PA}$}: Neural pair-wise: A FF layer between the concatenation of two hypothesis vectors and the nonlinear output layer. We run it $k$ - 1 times for a pair of hypotheses in a series of tournament-like competitions in the order of the $k$-best list. For instance, the 1st and 2nd hypothesis compete first and the winner of the two competes with the 3rd hypothesis next and so on until the $k$th hypothesis.

\begin{table}[t!]
\setlength\belowcaptionskip{-10pt}
\small
\centering
\begin{tabular}{l|ccc|cc}

            & \multicolumn{2}{c}{Traditional IPDA}    &     & \multicolumn{2}{c}{Large-Scale IPDA}             \\
            & \textit{$smx_a$}  & \textit{$smx_b$}     &     & \textit{$smx_a$}   & \textit{$smx_b$}     \\ \hline
1-best      & 95.58             & 95.56                &     & 81.38              & 81.49                \\
3-best      & 98.45             & 98.43                &     & 92.53              & 92.81                \\
5-best      & \textbf{98.81}    & \textbf{98.77}       &     & \textbf{95.77}     & \textbf{95.93}       \\ \hline

\end{tabular}
\caption{\small The $k$-best classification accuracies (\%) of Shortlister using different softmax functions in the traditional and large-scale IPDA settings.}
\label{tab:accuracy_base}
\end{table}

\item \textit{$N^{CH}$}: Neural quasi list-wise with manual cross-hypothesis features added to \textit{$N^{PO}$}, following past approaches~\cite{Robichaud2014,Crook2015,Khan2015} such as the ratio of Shortlister scores to the maximum score, relative number of slots across all hypotheses, etc.
\item \textit{$LSTM^{O}$}: Using only the BiLSTM output vectors as the input to the FF-layer.
\item \textit{$LSTM^{S}$}: Summing up the hypothesis vector and the BiLSTM output vectors as the input to the FF-layer, similar to residual networks \cite{He2016}.
\item \textit{$LSTM^{C}$}: Concatenating the hypothesis vector and the BiLSTM output vectors as the input to the FF-layer.
\item \textit{$LSTM^{CH}$}: Same as \textit{$LSTM^{C}$} except that manual cross-hypothesis features used for \textit{$N^{CH}$} were also added to see if combining manual and automatic cross-hypothesis features help.
\item \textit{$UPPER$}: Upper bound of HypRank accuracy set by the performance of Shortlister.
\end{itemize}

\subsection{Methodology}
Table \ref{tab:exam_nums} shows the distribution of the training, development and test sets for each setting of traditional and large-scale IPDAs. Note that we ensure no overlap between the Shortlister and HypRank training sets so that HypRank is not overly tuned on Shortlister results. For the NLU models, the intent and slot models are trained on roughly 70\% of the available training data.

In our experiments, all the models were implemented using Dynet~\cite{neubig2017dynet} and were trained with Adam~\cite{kingma2014adam}. We used the initial learning rate of 4 $\times 10^{-4}$ and left all the other hyper-parameters as suggested in ~\newcite{kingma2014adam}. 
We also used variational dropout~\cite{Gal2016} for regularization.

\begin{table}[t!]
\setlength\belowcaptionskip{-10pt}
\small
\centering
\begin{tabular}{l|ccc|cc}
            & \multicolumn{2}{c}{Traditional IPDA}        &     & \multicolumn{2}{c}{Large-Scale IPDA}                     \\
Model       & \textit{$smx_a$}      & \textit{$smx_b$}     &     & \textit{$smx_a$}       & \textit{$smx_b$}         \\ \hline
$SL$        & 95.58                 & 95.56                &     & 81.38                & 81.49                  \\ \hline
$LR$        & 95.50                 & 95.59                &     & 86.74                & 87.50                  \\
$N^{PO}$    & 95.26                 & 95.46                &     & 88.65                & 90.38                  \\
$N^{PA}$    & 96.08                 & 96.37                &     & 88.29                & 90.82                  \\
$N^{CH}$    & 96.20                 & 96.65                &     & 88.43                & 91.13                  \\
$LSTM^{O}$  & 94.36                 & 94.45                &     & 81.37                & 82.98                  \\
$LSTM^{S}$  & 97.44                 & 97.54                &     & 92.43                & 93.79                  \\
$LSTM^{C}$  & \textbf{97.47}        & \textbf{97.55}       &     & \textbf{92.49}       & \textbf{93.83}         \\
$LSTM^{CH}$  & 97.22                 & 97.34                &     & 92.18                & 93.46                  \\ \hline
$UPPER$     & 98.81                 & 98.77                &     & 95.77                & 95.93                  \\ \hline
\end{tabular}
\caption{\small The final classification accuracies (\%) for different Shortlisting-HypRank models under the traditional and large-scale IPDA settings. The input hypothesis size is 5.}
\label{tab:accuracy_full}
\end{table}

\subsection{Results and Discussion}

Table \ref{tab:accuracy_base} summarizes the $k$-best classification accuracy results for our Shortlister.
With only 20 domains in the traditional IPDA setting, the accuracy is over 95\% even when we take 1-best or top domain returned from Shortlister. The accuracy approaches 99\% when we consider Shortlister correct if the ground-truth domain is present in the top 5 domains. The results suggest that the character and word-level information by itself, coupled with BiLSTMs, can already show significant discriminative power for our task.

With a scale of 1,500 domains, the results indicate that just using the top domain returned from Shortlister is not enough to have comparable performance shown in the traditional IPDA setting. However, the performance catches up to close to 96\% as we include more domains in the $k$-best list, and although not shown here, it starts to level off at 5-best list. The $k$-best results from Shortlister set an upper bound for HypRank performance. We note that it could be possible to include more contextual information at the shortlisting stage to bring Shortlister's performance up with some trade-offs in terms of real-time systems, which we leave for future work. In addition, using \textit{$smx_b$} shows a  tendency of slightly better performance compared to using \textit{$smx_a$}, which takes a softmax over all domains and tends to emphasize only the top domain while suppressing all others even when there are many overlapping and very similar domains.

The classification performance after the reranking stage with HypRank using Shortlister's 5-best results is summarized in Table~\ref{tab:accuracy_full}. SL shows the results of taking the top domain from Shortlister without any reranking step, and UPPER shows the performance upper bound of HypRank set by the shortlisting stage.
In general, the pair-wise approach is shown to be better than the point-wise approaches, with the best performance coming from the list-wise ones. Looking at the lowest accuracy from \textit{$LSTM^O$}, it suggests that the raw hypothesis vectors themselves are important features that should be combined with the cross-hypothesis contextual features from the LSTM outputs for best results. Adding manual cross-hypothesis features to the automatic ones from the LSTM outputs do not improve the performance.

The performance trend is very similar for \textit{$smx_a$} and \textit{$smx_b$}, but there is a gap between them in the large-scale setting. An explanation for this is similar to that for Shortlister results that \textit{$smx_a$} emphasizes only the top domain while suppressing all the rest, which might not be suitable in a large-scale setting with many overlapping domains. For both traditional and large-scale settings,  the best accuracy is shown with the list-wise model of \textit{$LSTM^C$}.

\section{Conclusion}
We have described an efficient and scalable shortlisting-reranking neural models for large-scale domain classification. The models first efficiently prune all domains to only a small number of $k$ candidates using minimal information and subsequently rerank them using additional contextual information that could be more expensive in terms of computing resources. We have shown the effectiveness of our approach with 1,500 domains in a real IPDA system and evaluated  using different variations of the shortlisting model and our novel reranking models, in terms of point-wise, pair-wise, and list-wise ranking approaches.
\newpage
\section*{Acknowledgments}
We thank Sunghyun Park and Sungjin Lee for helpful discussion and feedback.
\bibliography{naaclhlt2018}
\bibliographystyle{acl_natbib}

\newpage
\appendix

\end{document}